\documentclass[nohyperref]{article}

\usepackage{microtype}
\usepackage{graphicx}
\usepackage{subfigure}
\usepackage{booktabs} 

\usepackage{hyperref}

\usepackage[dynnworkshop]{icml2022}

\usepackage{amsmath}
\usepackage{amssymb}
\usepackage{mathtools}
\usepackage{amsthm}

\usepackage[capitalize,noabbrev]{cleveref}

\theoremstyle{plain}

\theoremstyle{definition}

\theoremstyle{remark}

\usepackage[textsize=tiny]{todonotes}

\usepackage{amssymb}
\newcommand{\bb}[1]{\textcolor{blue}{#1}}

\usepackage{enumitem}

\usepackage{wrapfig}

\usepackage[T1]{fontenc}
\usepackage[utf8]{inputenc}
\usepackage{pmboxdraw}

\icmltitlerunning{Noisy Heuristics Neural Architecture Search}

\begin{document}

\twocolumn[
\icmltitle{Noisy Heuristics NAS: \\ A Network Morphism based Neural Architecture Search using Heuristics}



\icmlsetsymbol{equal}{*}

\begin{icmlauthorlist}
\icmlauthor{Suman Sapkota}{naamii}
\icmlauthor{Binod Bhattarai}{ucl}
\end{icmlauthorlist}

\icmlaffiliation{naamii}{NAAMII, Nepal}
\icmlaffiliation{ucl}{University College London, UK}

\icmlcorrespondingauthor{Suman Sapkota}{suman.sapkota@naamii.org.np}
\icmlcorrespondingauthor{Binod Bhattarai}{b.bhattarai@ucl.ac.uk}

\icmlkeywords{Machine Learning, ICML}

\vskip 0.3in
]



\printAffiliationsAndNotice{We would like to acknowledge Google Cloud for computing credits.}  

\begin{abstract}

Network Morphism based Neural Architecture Search (NAS) is one of the most efficient methods, however, knowing where and when to add new neurons or remove dis-functional ones is generally left to black-box Reinforcement Learning models. In this paper, we present a new Network Morphism based NAS called Noisy Heuristics NAS which uses heuristics learned from manually developing neural network models and inspired by biological neuronal dynamics. Firstly, we add new neurons randomly and prune away some to select only the best fitting neurons. Secondly, we control the number of layers in the network using the relationship of hidden units to the number of input-output connections. Our method can increase or decrease the capacity or non-linearity of models online which is specified with a few meta-parameters by the user. Our method generalizes both on toy datasets and on real-world data sets such as MNIST, CIFAR-10, and CIFAR-100. The performance is comparable to the hand-engineered architecture ResNet-18 with the similar parameters.


\end{abstract}

\section{Introduction and Related Works}

Neural Architecture Search (NAS) is the process of searching the Architecture of Neural Networks by leveraging the computation rather than doing manually. However, NAS has still not been able to come to the mainstream due to large computational costs and availability of more efficient alternatives such as transfer learning~\cite{zhuang2020comprehensive} or reusing architectures. Research has been done on using Reinforcement Learning(RL)~\cite{zoph2016neural, baker2016designing} and Genetic Algorithm(GA)~\cite{desell2017large} for generating architecture from a given search space, however, these methods have huge computational costs and produce large carbon footprints~\cite{strubell2019energy}. Gradient-based path selection methods such as DARTS~\cite{liu2018darts} and PC-DARTS~\cite{xu2019pc} have made NAS more efficient and accessible. However, it still involves training large parameter models and selecting only a subset for the final model.

In a manual architecture search process, we start with a baseline model architecture. If the model has poor performance, we test more and more non-linear models and if the model overfits the dataset we test smaller models, throwing away older models in the process. The initial concern is whether non-linear capacity can be increased or decreased in the same model while reusing the trained function. To our aid, Network Morphing based methods~\cite{elsken2017simple, lu2018compnet, dai2019nest, evci2022gradmax} have been used widely to add neurons, which increase the non-linearity and capacity of the model.

However, Network Morphism based methods are generally paired up with Reinforcement Learning (RL)~\cite{cai2018efficient} or Bayesian Optimization~\cite{jin2019auto}, which decide the morphism operation to increase the network capacity. This type of solution makes the dynamic nature of neural network a difficult to understand. To understand the dynamics and to simulate heuristics, we need easily controllable models for changing network capacity or the number of neurons or parameters. 

Although there are various works on using Network Morphism for Neural Architecture Search, we find that the methods are partial, either only adding neurons~\cite{jin2019auto, cai2018efficient} and layers or not pruning layers~\cite{gordon2018morphnet} to reduce capacity. Furthermore, those methods that add and prune neurons use it on incremental~\cite{dai2020incremental} or continual~\cite{zhang2020regularize} learning settings. Our goal to search for architecture depending on the dynamics requires additional components to change the structure (layers and neurons) of the network itself. This gap motivates us to create a Network Morphism based NAS that can change the number of layers and neurons dynamically during the training phase while keeping the search efficient and simple for the user. We combine multiple ideas and heuristics for creating a framework of Noisy NAS to search for capacity.



Ideas from pruning and dropout support our framework for noisy heuristic-based architecture search. When small neurons are pruned, they typically recover the same accuracy and loss~\cite{molchanov2019importance} without recovering the function completely. Furthermore, noisy regularization methods like Dropout~\cite{srivastava2014dropout} and DropConnect~\cite{wan2013regularization} suggest that Deep Networks can be trained to be robust to perturbations. We can infer that Neural Networks are robust to the noisy process of addition and pruning of neurons. We can use such a noisy process to try different additions and removals of neurons iteratively which can  roughly change the architecture to the desired capacity. The method of adding many neurons and removing poorly performing ones could be used to search for the correct place to add new neurons.

Furthermore, the dynamics of Biological Neural Networks (BNN)~\cite{wan2019single} suggests that there could exist Neural Networks with dynamically changing architecture in a single model. The dynamic nature of BNN is partly due to neurogenesis~\cite{kumar2019adult}, neuron and synaptic pruning~\cite{fricker2018neuronal} and neuron migration. We aim to understand the internal workings of Artificial Neural Networks(ANN) and apply dynamics from BNN to close the gap between them. We believe that Dynamic Neural Networks along with Spiking Neural Networks~\cite{tavanaei2019deep} could model BNN even better.





\begin{figure}
     \centering
     \includegraphics[trim= 0cm 0cm 0cm 0cm, clip, width=0.9\linewidth]{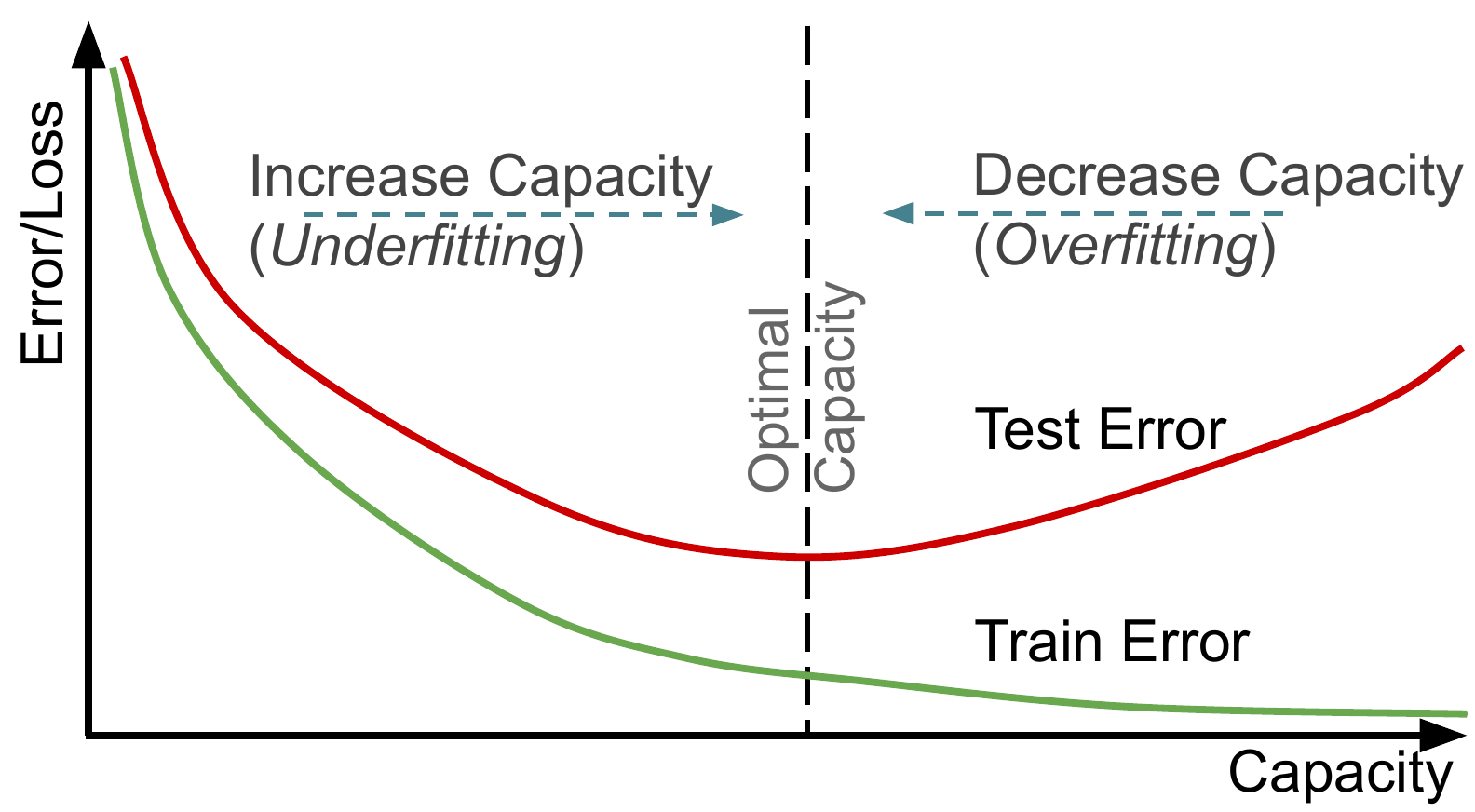}
     \caption{Generalization Curve}
     \label{fig:generalization-curve}
\end{figure}

\textbf{Our Contribution: }
Combining the growing and shrinking mechanisms, we are able to get any desired network capacity for the best fitting of the dataset. Such a method is depicted by a generalization curve as shown in Figure~\ref{fig:generalization-curve}. We work on the same curve, but instead of trying different capacity models, we change the capacity of the existing models towards the best capacity.
To this end, we propose a new method for Network Morphism based Neural Architecture search using heuristics. We simplify our search space using multiple heuristics to a manageable number of meta-parameters. The major contributions of our work are listed below.
\begin{enumerate}[noitemsep, topsep=0pt]
    \item We introduce a new method to add new neurons and layers heuristically for Network Morphism based Neural Architecture Search.
    \item We create a new type of architecture called Hierarchical Residual Network for the ease of changing non-linearity and number of layers during Network Morphism.
    \item We combine neuron addition, pruning and Hierarchical Residual Network to change the capacity of the network noisily during training, which we call Noisy Heuristics NAS.
    \item We show that our method is successful in getting performance near hand-designed architectures like ResNet.
    \item We release code for Network Morphism, Optimizer reusing, Pruning and Noisy Heuristic NAS in the PyTorch~\cite{paszke2019pytorch} framework. \small{\bb{https://github.com/tsumansapkota/Noisy-Heuristics-NAS}}
\end{enumerate}

\section{Methodology}

The following concepts are combined to create a Noisy Neural Architecture Search based on Heuristics.

\textbf{Hiearchical Residual Network } We propose a generalization of ResNet~\cite{he2016deep} called Hierarchical Residual Network (H-ResNet). The main concept is that each linear layer can be made non-linear by adding a residual function to it, which is similar to ResNet, as shown by equation~(\ref{eqn:1}). Such residual connections are easy to add and remove without breaking the flow of information in Deep Neural Networks. We extend this idea to the limit. The addition of a residual function to the linear connection can be extended to the linear connection of the residual connection itself as shown by equation~(\ref{eqn:2}). 
\vspace{-7pt}
\begin{equation}
  f(x) = W(x) + Re(x) \text{\space ; \space} Re(x) = W_2(\sigma(W_1(x))
  \label{eqn:1}
\end{equation}
\vspace{-20pt}
\begin{equation}
  Re(x) = f_2(\sigma(f_1(x)))
  \label{eqn:2}
\end{equation}
\vspace{-20pt}

\begin{figure}
     \center
     \includegraphics[trim= 0cm 0cm 0cm 0cm, clip, width=0.9\linewidth]{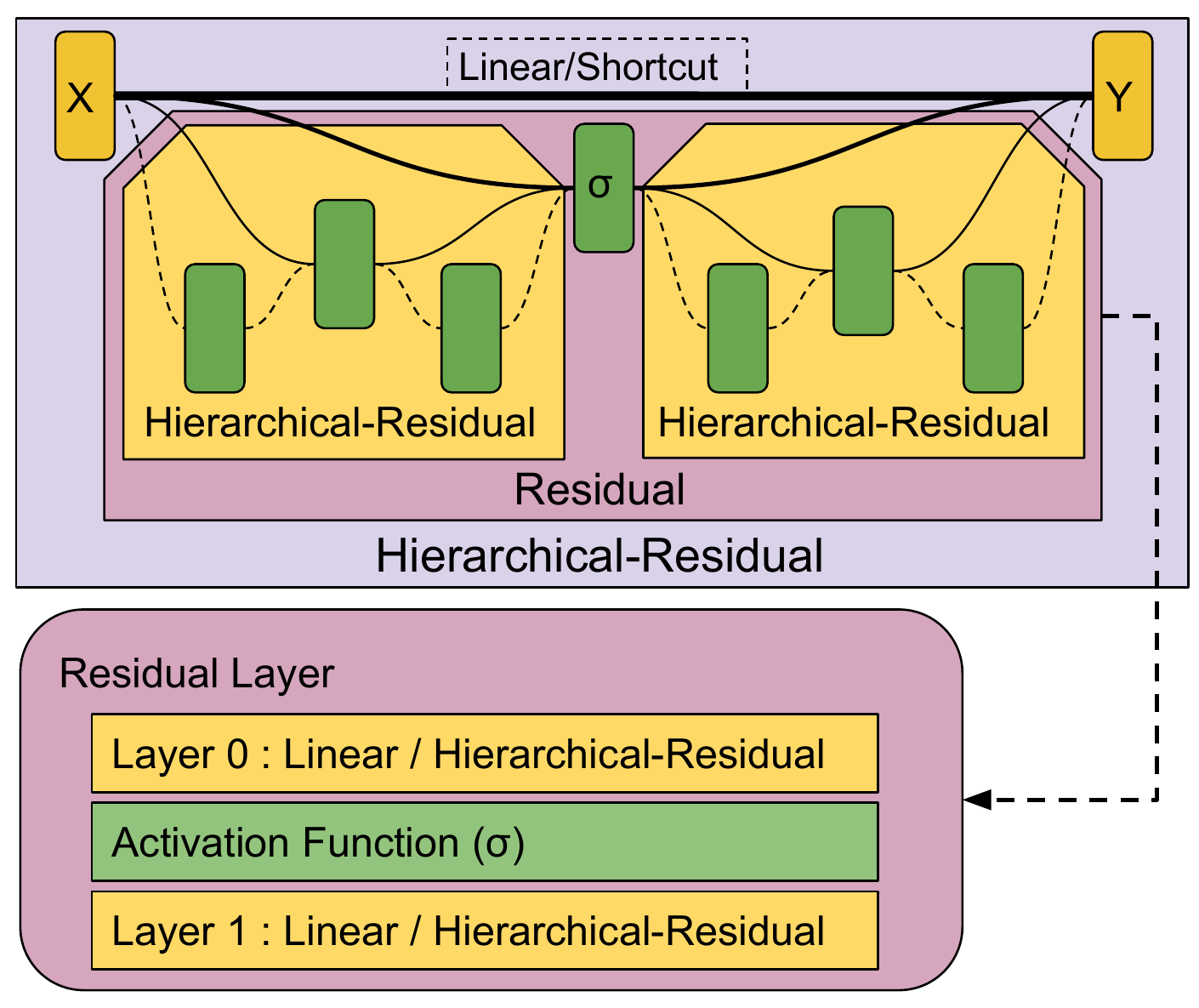}
     \caption{Hierarchical Residual Network}
     \label{fig:h-resnet}
\end{figure}

Here, $f_1$ and $f_2$ are ResNets with different parameter. The hierarchical nature of this type of network is depicted by the figure~\ref{fig:h-resnet}. Theoretically, all Ordinary Networks, ResNets~\cite{he2016deep} and UNets~\cite{ronneberger2015u} are special cases of Hierarchical ResNets. If the shortcut connection is zero then it is an Ordinary Network and if the residual function has shortcut as its layers then it is a ResNet. 
 
\textbf{Where to add new neurons ? }
The problem of where to add new neurons and how many has large possible choices. To tackle this, we first add large number of neurons ($P$) distributed to all layers. Secondly, we train the model to fit the new neurons to data. Thirdly, we prune $M$ least important neurons such that change in number of neurons is given by, $\Delta N = P-M$.In this way, we are able to find a good solution for where to add new neurons. 

The change in number of neurons in each layer is given by, $\Delta N_i = P_i - M_i$, where $P_i$ is the initial number of neurons and $M_i$ is the number of pruned neurons. To accelerate the addition of neurons in correct layers in next iterations, we add $70\%$ of new neurons with the probability of each layer $p_i \propto \text{MA}(\Delta N_i)$, where MA represents Moving Average function change in number of neurons.


\textbf{Where to add new layers ?}
In Hierarchical ResNet, we can add layers to any linear connection of any hierarchy. To solve the problem of having a large number of possible choices, we use a rough heuristic to add new layers to \textit{Residual Layer} automatically and heuristically after \textit{Shrink Phase}. The heuristic is that if hidden neurons ($H$) on Residual Layer with Linear connection is greater than $M = (I*O^2)^\frac{1}{3}$, we convert the inner layers, \textit{Layer 0} and \textit{Layer 1}, from Linear to Hierarchical-Residual Layer. The Hierarchical-Residual Layer can again grow layers on its Residual Layer in a recursive way. (See Figure~\ref{fig:h-resnet}) 





\textbf{How to prune neurons ?}
We use global importance estimation based pruning similar to previous methods~\cite{molchanov2019importance, yu2018nisp} which prunes the least important neurons. The importance score per neuron ($I$) is given by 
$I = A*(1-B^{33})$. Where,
\vspace{-7pt}
$$A = \Sigma_{i=0}^N{|
  \sigma_i * \delta \sigma _i|}
  \text{\space ; \space}
  B = \frac{1}{N}\Sigma_{i=0}^N(1[\sigma_i > 0] )$$
\vspace{-20pt}

$\sigma_i$ is the activation for $i^{th}$ data point, $\delta \sigma _i$ is the gradient of the activation and $N$ is the total number of data in the dataset. The term $B$ gives fraction of non-zero activations. We scale the term $A$ by $B$ to give low importance to always firing neurons.

Since pruning removes non-zero neurons as well it creates sudden change in function learned. To make removal gradual, we decay the outgoing connection of least important neurons while finetuning other neurons for some epochs.





\textbf{How to remove layers ?}
During each iteration of addition and pruning, we morph the network at the end of the pruning stage. After pruning, if the number of neurons in a certain Residual Layer (see Figure~\ref{fig:h-resnet}) decreases to near zero (say 1), then we prune away the neurons from the layer and the layer is removed altogether by our algorithm. This makes the parent Residual layer contain Linear connection as \textit{Layer 0} or \textit{Layer 1}. This happens just the opposite way of adding a new Layer. New layers added in the previous iteration are susceptible to removal.

\textbf{When to know when model is fitted ?}
In a general training setting, we stop the training when the loss curve flattens out. 
First, we normalize steps and loss values to range [0, 1] and fit the steps vs loss data to $y=e^{ax}(1-x)$, where $a$ is the parameter (See Appendix~\ref{sec:loss_curve_flattening}). We set $a<-5$ as threshold for determining the flattening of curve. The training is stopped if $a<-5$ or epochs $\ge$ maximum train epochs.  




\paragraph{Noisy Heuristic NAS}

\begin{figure}
     \center
     \includegraphics[trim= 0cm 0cm 0cm 0cm, clip, width=0.98\linewidth]{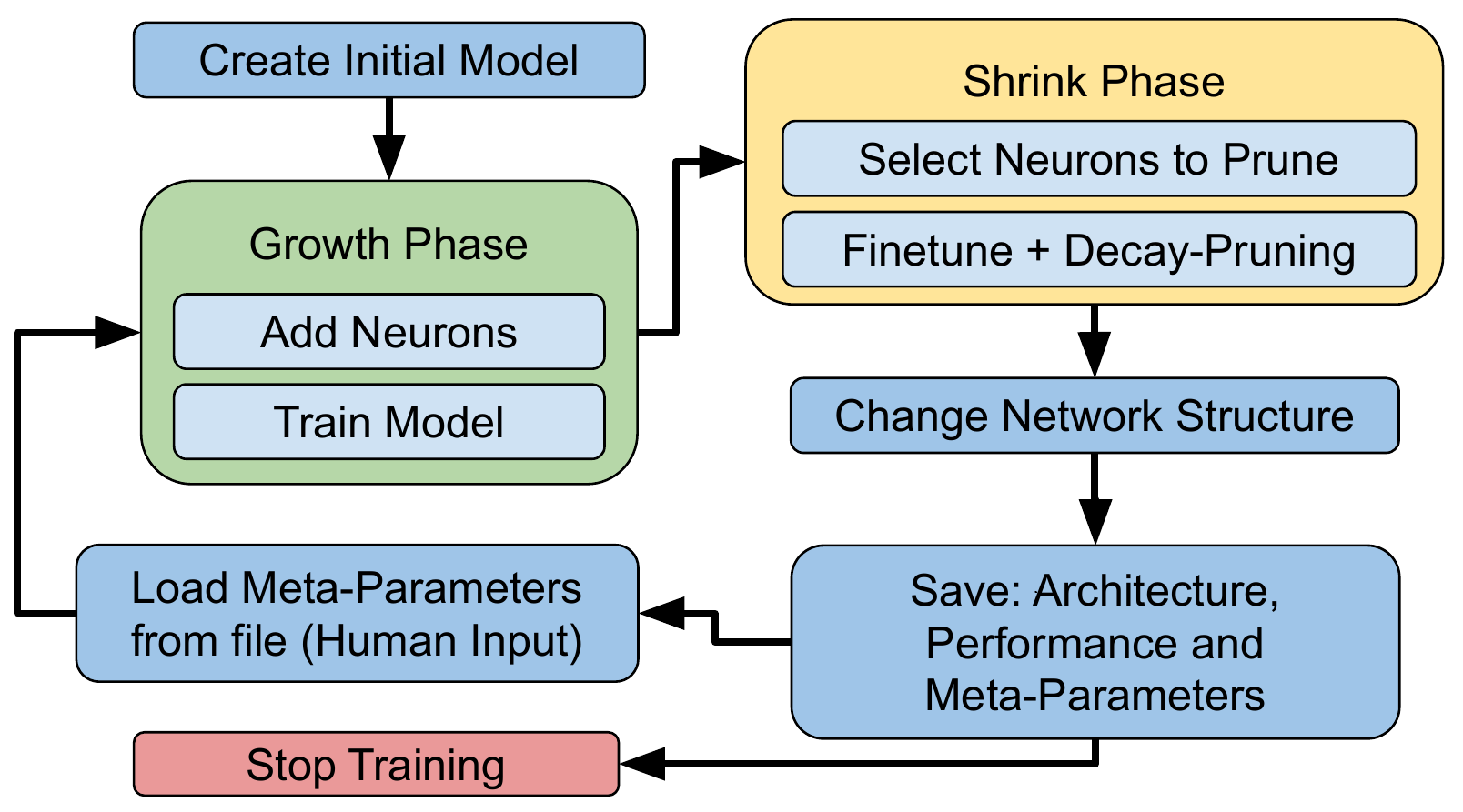}
     \caption{Pipeline of Noisy Heuristic NAS}
     \label{fig:noisy-pipeline}
\end{figure}

Our method combines the above-mentioned operations to search for the architecture. The pipeline for the Noisy Heuristic NAS is shown in Figure~\ref{fig:noisy-pipeline}. The pipeline is explained as follows.

Firstly, an initial model is created. The initial model can be Linear Layer or some standard architecture like ResNet. Secondly, during the \textit{Growth Phase}, we add 
neurons randomly.
This model is then trained for given epoch or until convergence. Thirdly, during the \textit{Shrink Phase}, neurons are pruned by importance score. 


Next, we save the model, performance and current meta-parameters. Then we take meta-parameters as input from a file, which is an interface between user and the NAS system. We repeat the process of growth, shrink, save and load for multiple iterations until desired network is achieved.


\begin{figure*}[t]
  \centering
  \subfigure{\includegraphics[trim= 0.1cm 0.2cm 0.8cm 1cm, clip, width=0.29\textwidth]{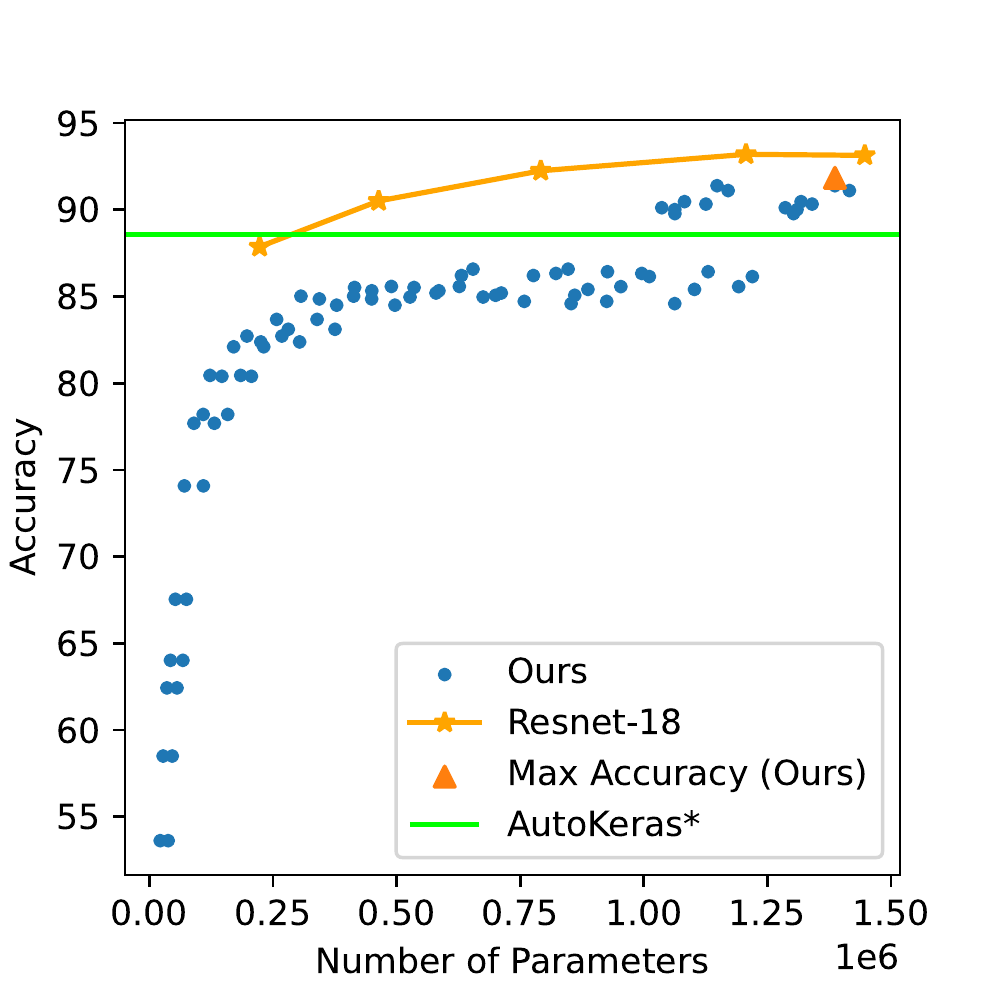}}
  \subfigure{\includegraphics[trim= 0.1cm 0.2cm 0.8cm 1cm, clip, width=0.29\textwidth]{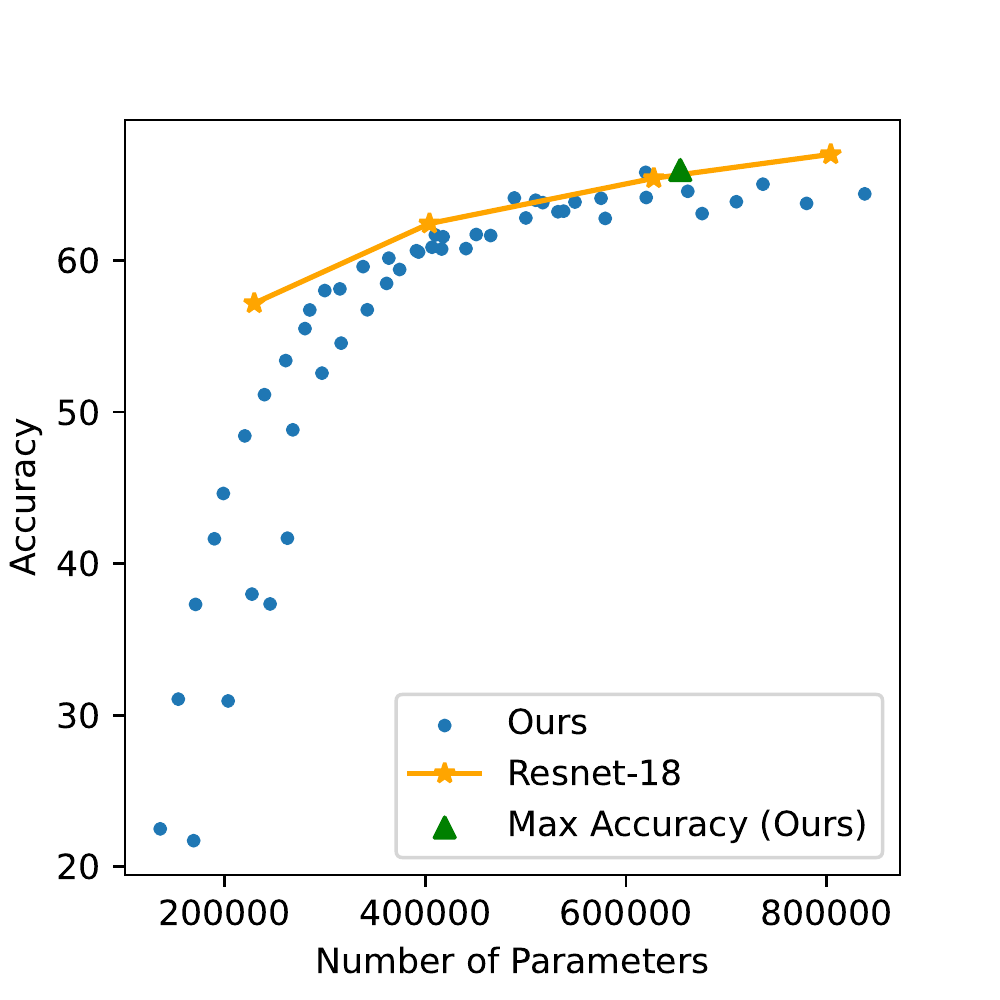}}
  \subfigure{\includegraphics[trim= 0.3cm 0.2cm 0cm 0.5cm, clip, width=0.38\textwidth]{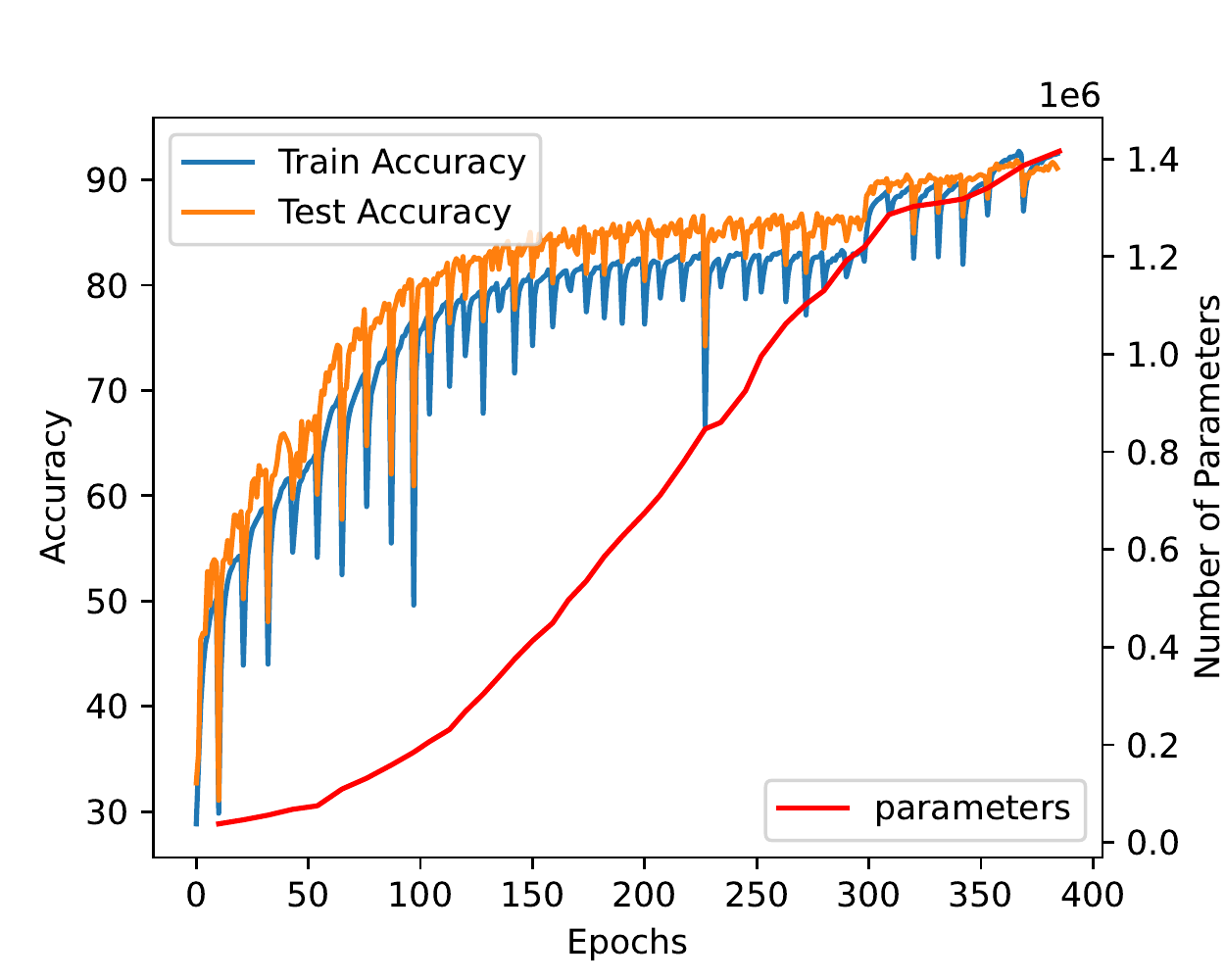}}
  \caption{
  \textbf{Left:} Parameter vs Accuracy plot for CIFAR-10 dataset. AutoKeras~\cite{jin2019auto} has unknown number of parameters. \textbf{Center:} Parameter vs Accuracy plot for CIFAR-100 dataset. \textbf{Right:} Epoch vs (Accuracy and Parameter) plot for CIFAR-10 dataset. 
  }
  \label{fig:experiment-cifar}
\end{figure*}

\textbf{Expanding or shrinking :} Our method allows the network to either expand or shrink in capacity allowing us to navigate the capacity dimension of Generalization Curve as shown in Figure~\ref{fig:generalization-curve}. If the number of neurons added is more than pruned, then the network is expanding, otherwise the network is shrinking. This allows us to change the capacity dynamically. We may start from pre-trained large network and decrease its capacity or start from linear model and increase its capacity towards large model all in a single continuous training, without throwing away models completely.







\textbf{Workflow: }The typical workflow of Noisy Heuristics NAS is to observe the performance and capacity and tweak the meta-parameters during the training. The major meta-parameters that are tuned manually during the search process are (1) number of neurons to add, (2) number of neurons to remove / add-to-remove-ratio and (3) learning rate.  


Using these parameters we can control the type of growth of the network. The architecture found is dependent on the dynamics of the process and the meta-parameters.

\textbf{Genetic Algorithm View of Noisy Heuristic NAS :}
We find our addition and pruning based approach of search
similar to Genetic Algorithm. If we view individual neurons as an organism, the pruning is equivalent to the selection process and new neurons are equivalent to new generation/mutation. Furthermore, we could apply better genetic algorithms for better selection of neurons and faster convergence of the search process.  

\section{Experiments}
\label{sec:experiment}

\textbf{Experiments on Toy Datasets:} 
We experiment on small 2D toy regression and classification problems to verify our method (See Appendix~\ref{sec:appendix_toy_data}). We find that our method effectively finds a suitable network for fitting the given dataset. We do not compare our method to any other architecture due to the simple solution to these datasets. However, it opens our method to be tested on large scale datasets like MNIST, CIFAR-10 and CIFAR-100. 


\textbf{Experiments on Large Scale Datasets :}
We find that our method improves the architecture search by modifying the network without changing the function. The use of heuristics for network morphism helps to avoid the computational cost of training deep RL models. 
We extend our method to MNIST dataset with Dense and Convolutional Layers where we get an accuracy of 97.87\% and 99.34\% respectively. We find that our method produces satisfactory results. 

Since our code is written mostly on Python/PyTorch and Hierarchical ResNet is not optimized, we find that it takes a longer time. We instead compare the methods in terms of the number of epochs for optimization. We find that our method performs comparatively with hand-designed ResNet-18 architecture on both CIFAR-10 and CIFAR-100 datasets (see Figure~\ref{fig:experiment-cifar}). Our method gradually changes (increases) parameters dynamically during the training which results in a gradual increase in test accuracy. A similar trend follows in the CIFAR-100 experiment as well.

For ResNet-18 architecture, we change the number of parameters by changing the width (number of channels) of the residual blocks. 
We choose ResNet with similar number of parameters for relative comparison.
The experiments are carried out for 200 epochs each. And for our method, meta-parameters were changed during the training to search desired capacity and for better performance.  

While testing for 5 different parameter ResNets, we spend a total of 1000 epochs which is higher than we train our method for ($\approx$ 400). Our method produces models of varying size and similar performance in one training. Further details of the experiments and findings are in Appendix~\ref{sec:appendix_experiment}.

\section{Conclusion}

In this work, we presented Noisy Heuristics NAS working on the Network Morphism framework. We efficiently train variable-capacity models by morphing the same network with minimal loss in trained parameters. Our method has performance comparable to hand-designed architecture like ResNets while allowing the network to change in capacity by meta-parameters adjusted during training.

\nocite{langley00}

\bibliography{biblography}
\bibliographystyle{icml2022}


\newpage
\appendix
\onecolumn



\section{Stopping Criterion using Loss Curve}
\label{sec:loss_curve_flattening}

\begin{figure}
     \centering
     \includegraphics[trim= 0cm 0cm 0cm 0cm, clip, width=0.6\linewidth]{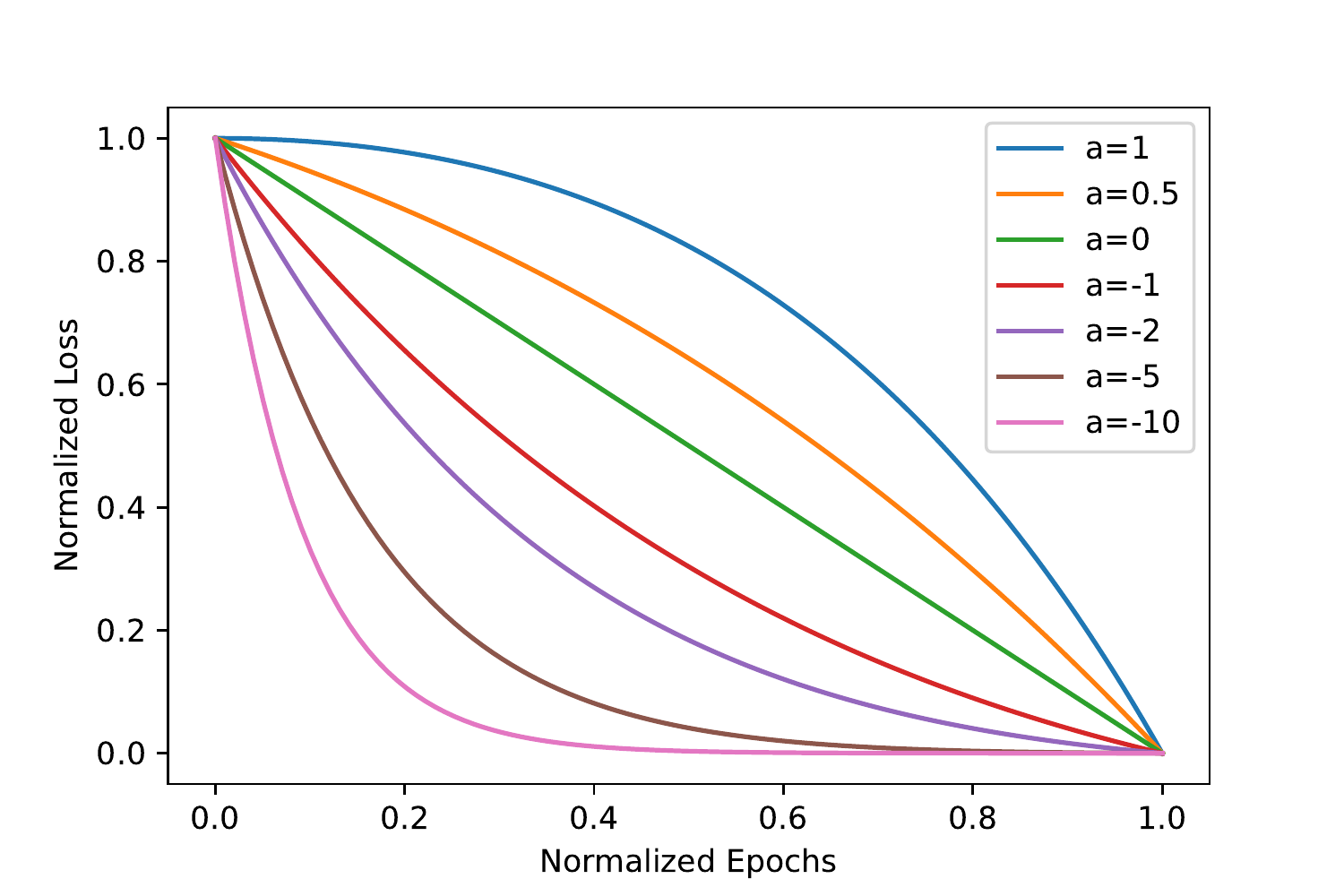}
     \caption{Model for flattening of loss curve with different parameters. Here the number of epochs and loss values are normalized in range [0,1] to fit the model.}
     \label{fig:loss-curve}
\end{figure}

We use the model $y=e^{ax}(1-x)$ for detecting the flattening of the loss curve. The Figure~\ref{fig:loss-curve} shows the model for different values of $a$. We set $a<-5$ for determining that the loss curve has flattened. We also set maximum training epochs threshold for stopping if curve does not flatten out. The maximum training epochs is also a meta-parameter which can be changed by the user. The experiments also show changing maximum training epochs change during the experiment.

\section{Toy Experiments}
\label{sec:appendix_toy_data}

\begin{figure}
  \centering
  \subfigure{\includegraphics[trim= 0.0cm 0.0cm 0.0cm 0.0cm, clip, width=0.35\textwidth]{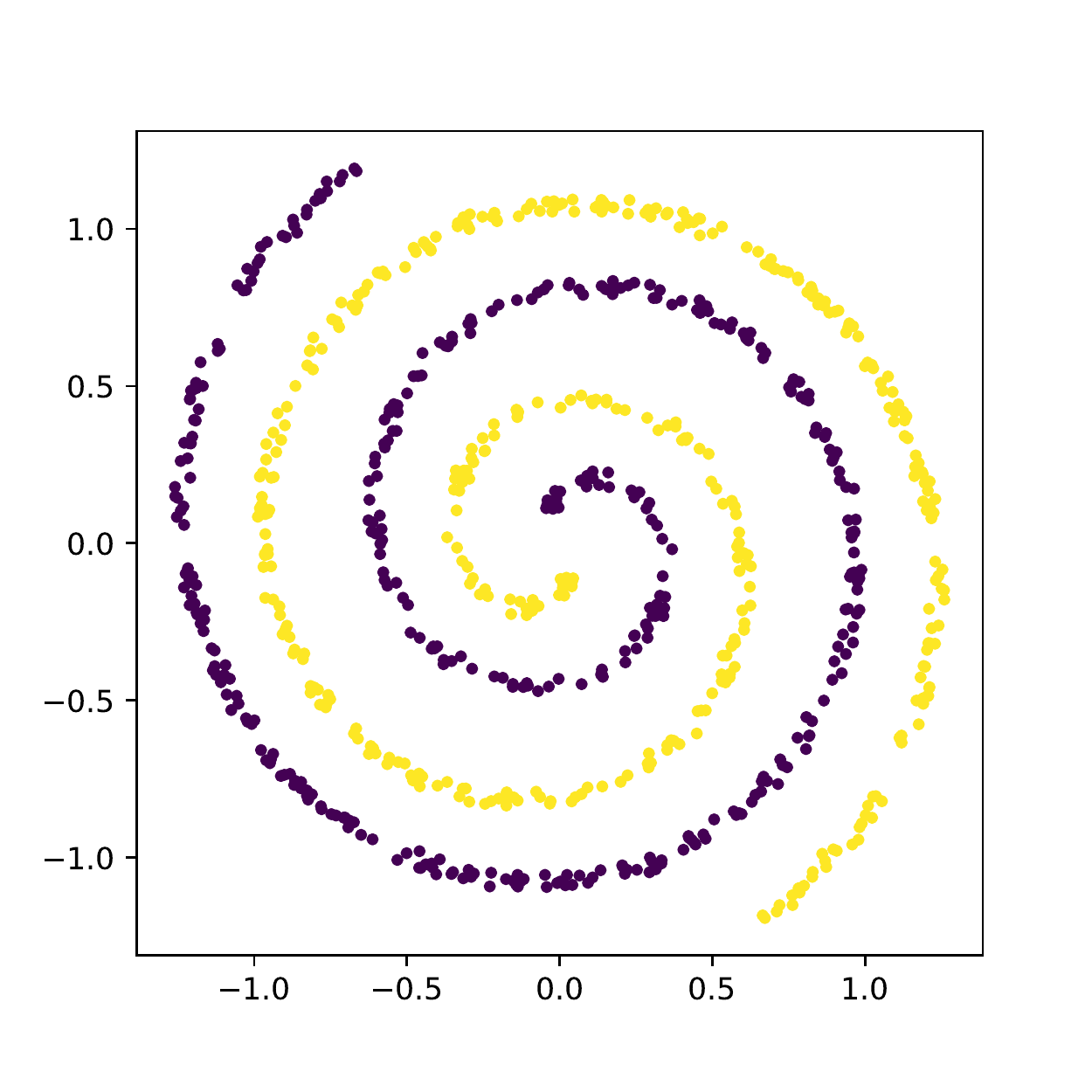}}
  \subfigure{\includegraphics[trim= 0.0cm 0.0cm 0.0cm 0.0cm, clip, width=0.35\textwidth]{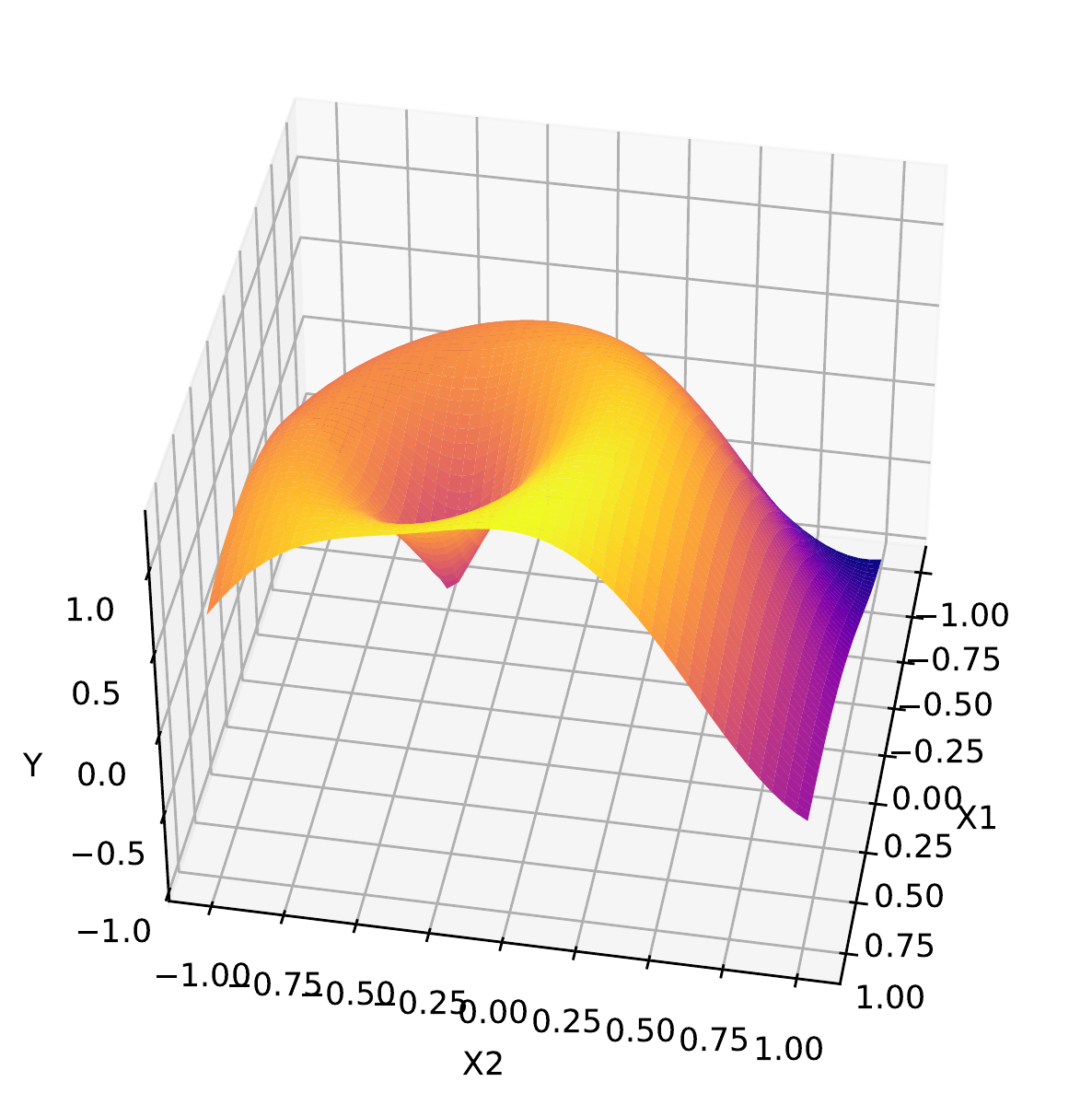}}
  
  \caption{
  \textbf{Left:} 2-Spiral Dataset for binary classification. \textbf{Right:} A regression dataset on a 2D grid.
  }
  \label{fig:toy_datasets}
\end{figure}

We use 2D synthetic datasets for verifying that our algorithm works. The spiral dataset shown in Figure~\ref{fig:toy_datasets} is used for classification. In the experiment, we get accuracy of 100\% in few iterations. Furthermore, we also test our algorithm on 2D regression dataset as shown in Figure~\ref{fig:toy_datasets}. We find that our method produces neural network that fits the dataset well.

\section{CIFAR Experiments: Extended}
\label{sec:appendix_experiment}

The experiments carried out in the Experiment Section (\ref{sec:experiment}) are extended and elaborated in this section. The experiments for ResNet-18 architectures on both CIFAR-10 and CIFAR-100 datasets are carried out with Adam optimizer with a learning rate of 0.001 with Cosine Scheduler with a 200 time period. For Noisy Heuristics NAS, we use Adam optimizer with learning rates manually changed during the training. We use dropout with drop probability $p=0.1$ with ReLU activation function. The use of dropout causes the Train Accuracy to be lower than Test Accuracy.

The meta-parameters were changed during the training to search desired capacity and for better performance. The exact meta-parameter values and observations are elaborated next.  

\paragraph{CIFAR-10 Experiment: }

\begin{figure}
     \centering
     \includegraphics[trim= 0.0cm 0.0cm 0.0cm 0.0cm, clip, width=1.0\textwidth]{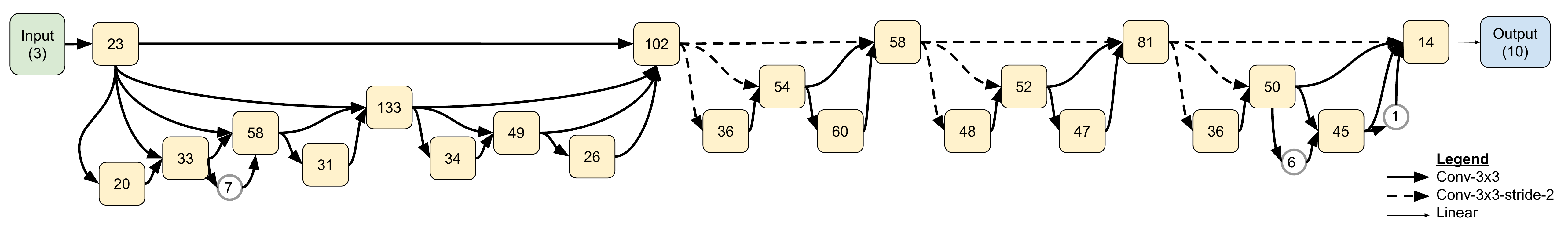}
     \caption{Architecture Found in CIFAR-10 experiment. The backbone network is inspired by ResNet architecture. We consider each block as Hierarchical Residual Network which is allowed to expand. The depth of this architecture is 23 (not counting the layers with the number of neurons < 10). We apply Global Average Pooling (GAP) before the Linear Layer for Classification.}
     \label{fig:architecture_found_c10}
\end{figure}

\begin{figure}
     \centering
     \includegraphics[trim= 0cm 0cm 0cm 0cm, clip, width=0.9\linewidth]{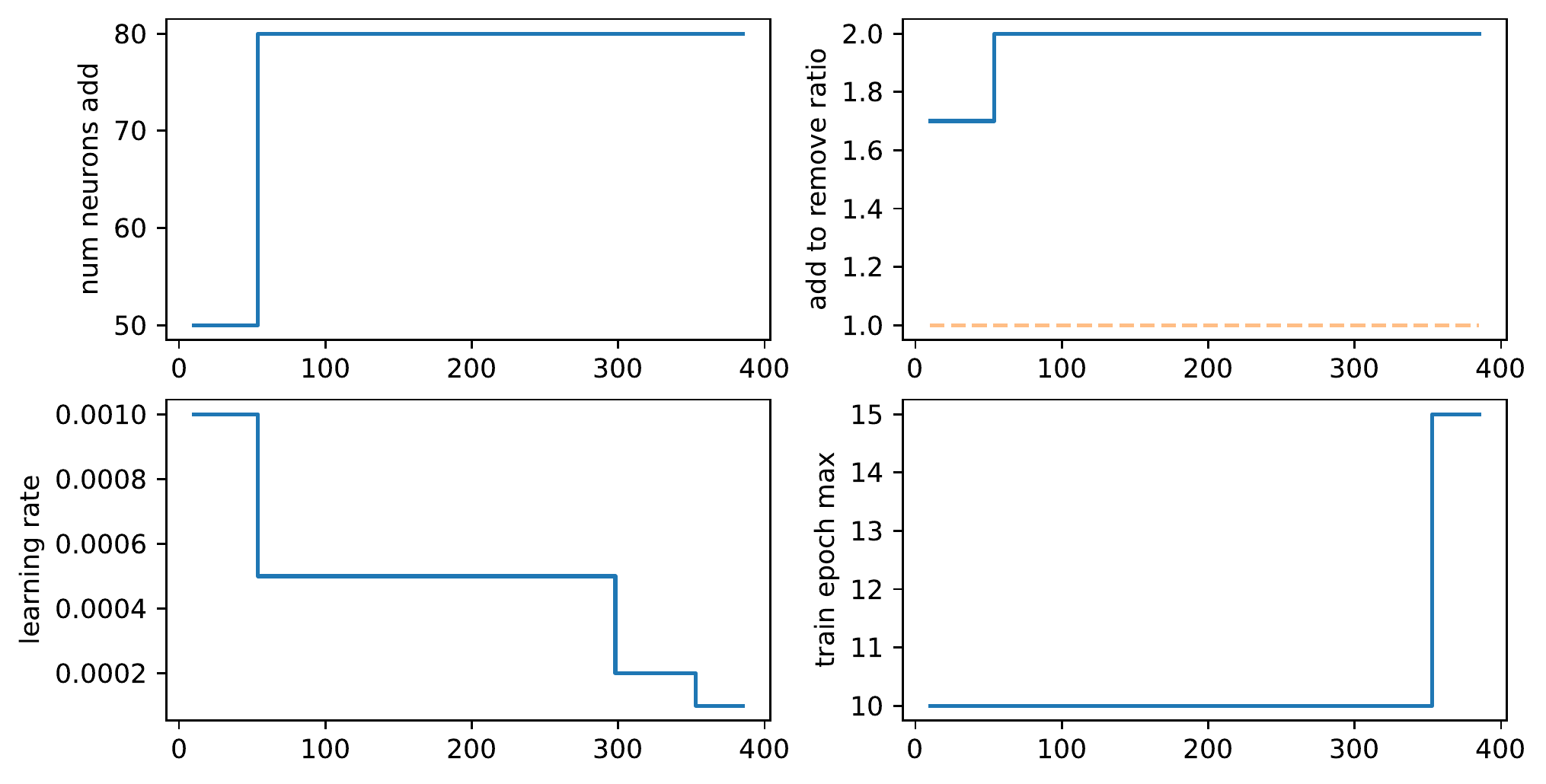}
     \caption{Different Meta-Parameters changed during the training phase. The x-axis is the number of epochs and y-axis (labeled) are meta-parameters}
     \label{fig:meta_param_c10_v0}
\end{figure}

Firstly, experiment on the CIFAR-10 dataset using our method, as shown in the Experiment Section produces best architecture with $1.42$M parameters and $91.85$\% accuracy. Furthermore, Residual Network produces $93.14$\% accuracy with $1.45$ parameters. The accuracy mentioned in Auto-keras paper~\cite{jin2019auto} is $88.56$\% with an unknown number of parameters. 
The CIFAR-10 experiment carried out in the Experiment Section has varying meta-parameters in different epochs. We plot the meta-parameters in the Figure~\ref{fig:meta_param_c10_v0}. Furthermore, we plot the architecture found using our search method in its best epoch in Figure~\ref{fig:architecture_found_c10}

\begin{figure}
  \centering
  \subfigure{\includegraphics[trim= 0.1cm 0.2cm 0.8cm 1cm, clip, width=0.35\textwidth]{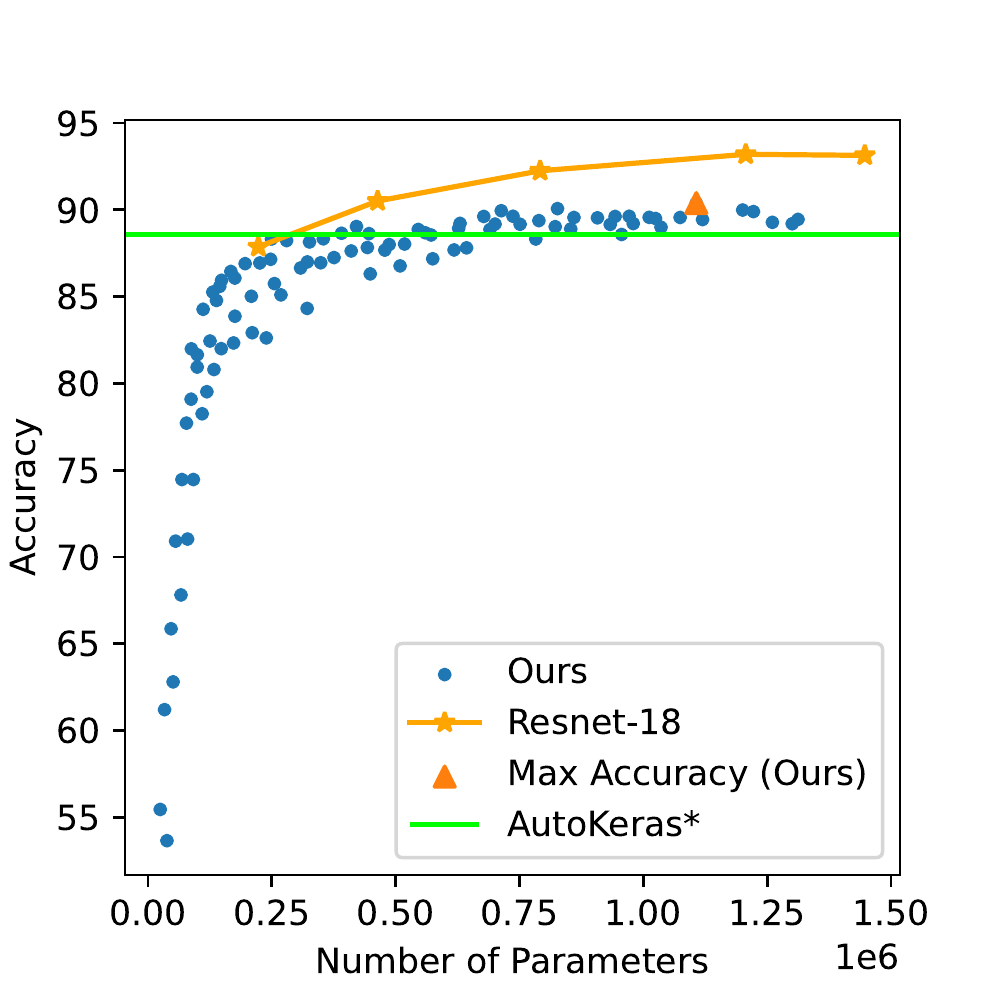}}
  \subfigure{\includegraphics[trim= 0.3cm 0.2cm 0cm 0.5cm, clip, width=0.45\textwidth]{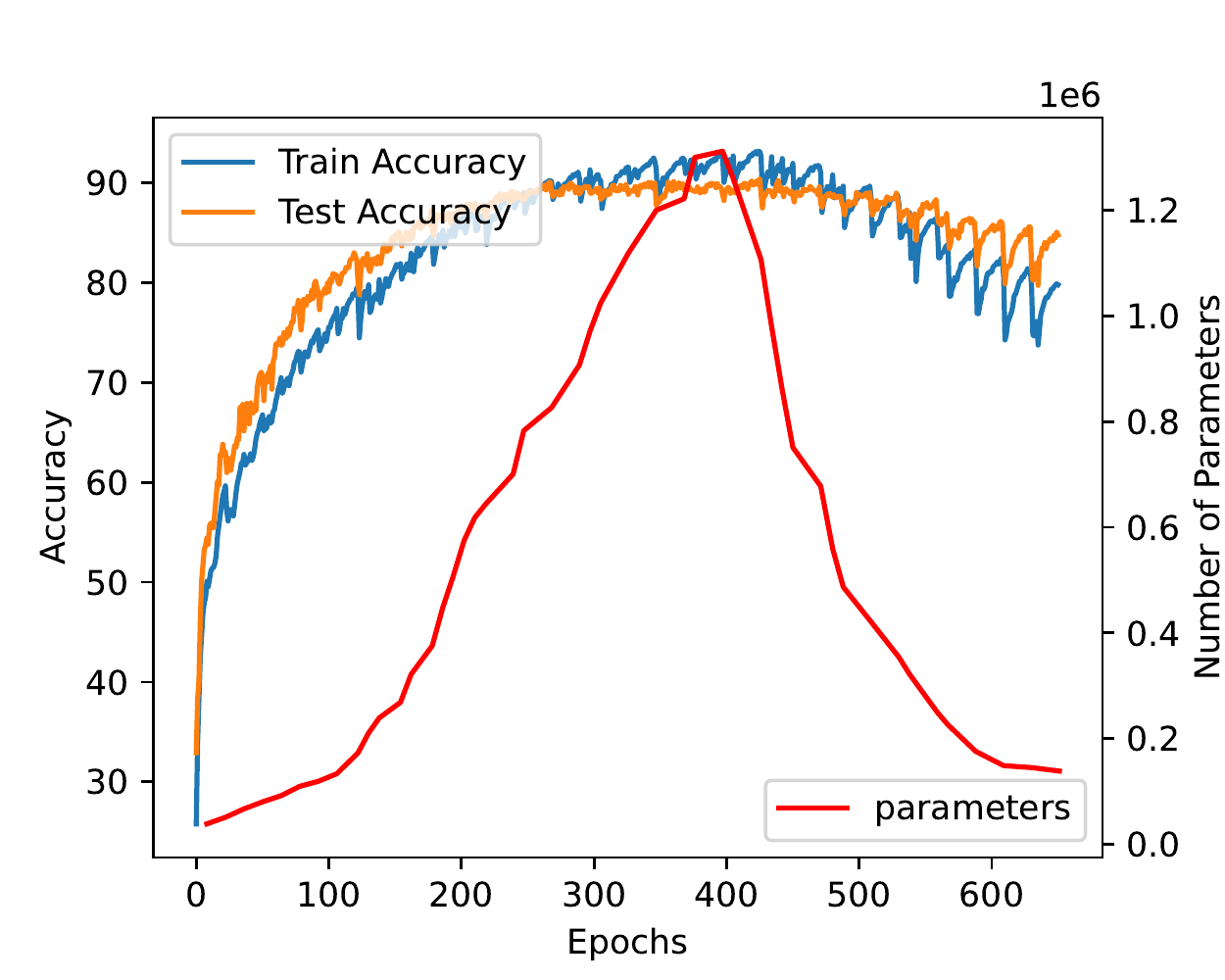}}
  \subfigure{\includegraphics[trim= 0.0cm 0.0cm 0cm 0.0cm, clip, width=0.8\textwidth]{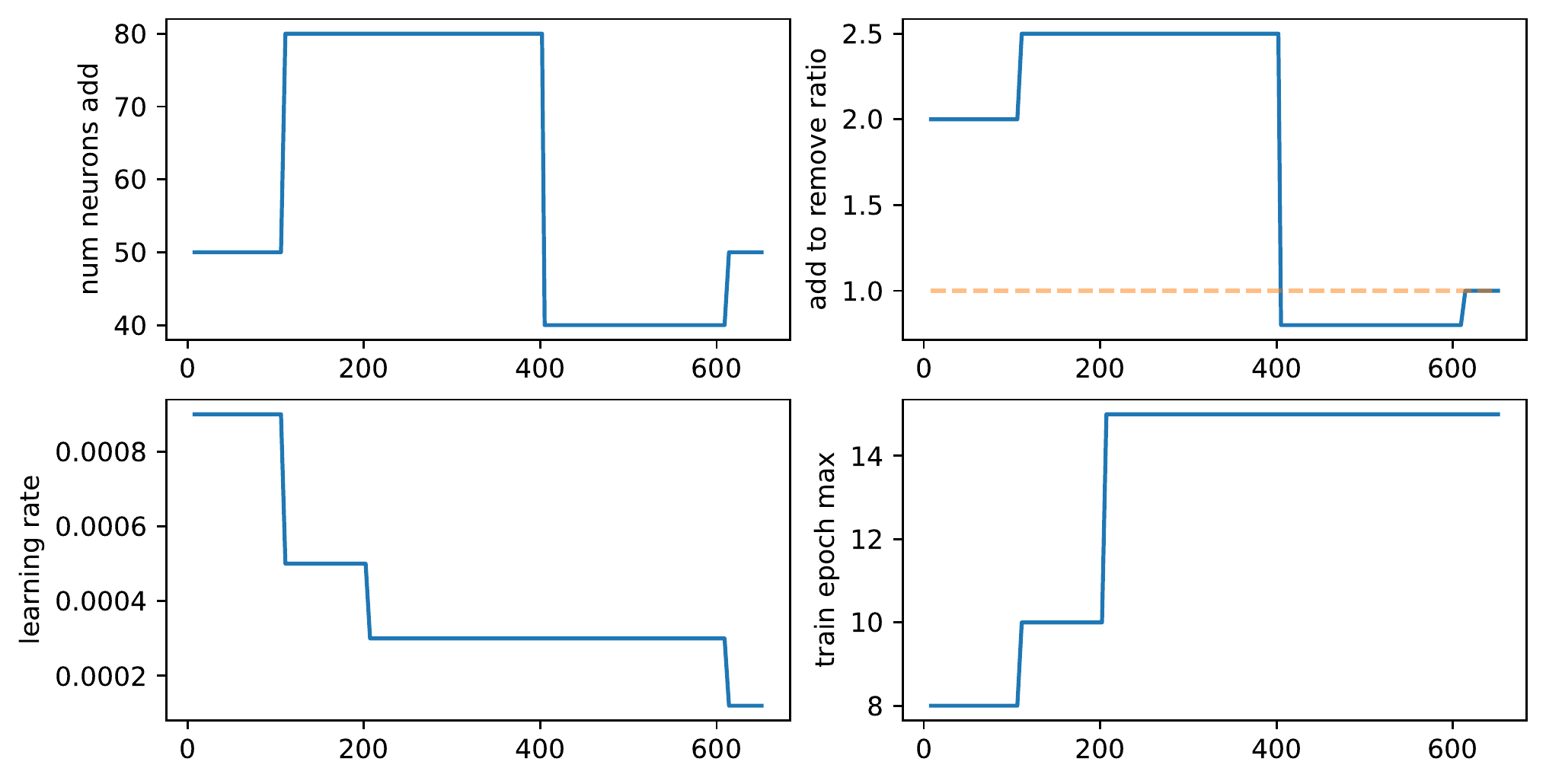}}
  
  \caption{
  \textbf{Top Left:} Parameter vs Accuracy plot for CIFAR-10 dataset. \textbf{Top Right:} Epoch vs (Accuracy and Parameter) plot for CIFAR-10 dataset. \textbf{Bottom Four:} Meta-parameters that guide the architecture search process.
  }
  \label{fig:experiment-cifar-10_v1}
\end{figure}

Furthermore, to verify that our algorithm can also reduce the parameters from a large model, we first grow the model (till epoch 400) and later shrink the model. The observations from this experiment are shown in Figure~\ref{fig:experiment-cifar-10_v1}. It verifies that our algorithm can effectively reduce neurons and layers according to the shrinking or expanding nature of the search process. The architecture also has a varying number of depths while increasing or decreasing the number of neurons. In the initial stage, the depth of the network is 10, at peak accuracy, it is 26 and at the later stage of shrinking, the depth is 16. Furthermore, it can be observed that the accuracy is correlated with the number of parameters.

\clearpage
\paragraph{CIFAR-100 Experiment:}

\begin{figure}
     \centering
     \subfigure{
     \includegraphics[trim= 0.0cm 0.0cm 0.0cm 0.0cm, clip, width=0.5\textwidth]{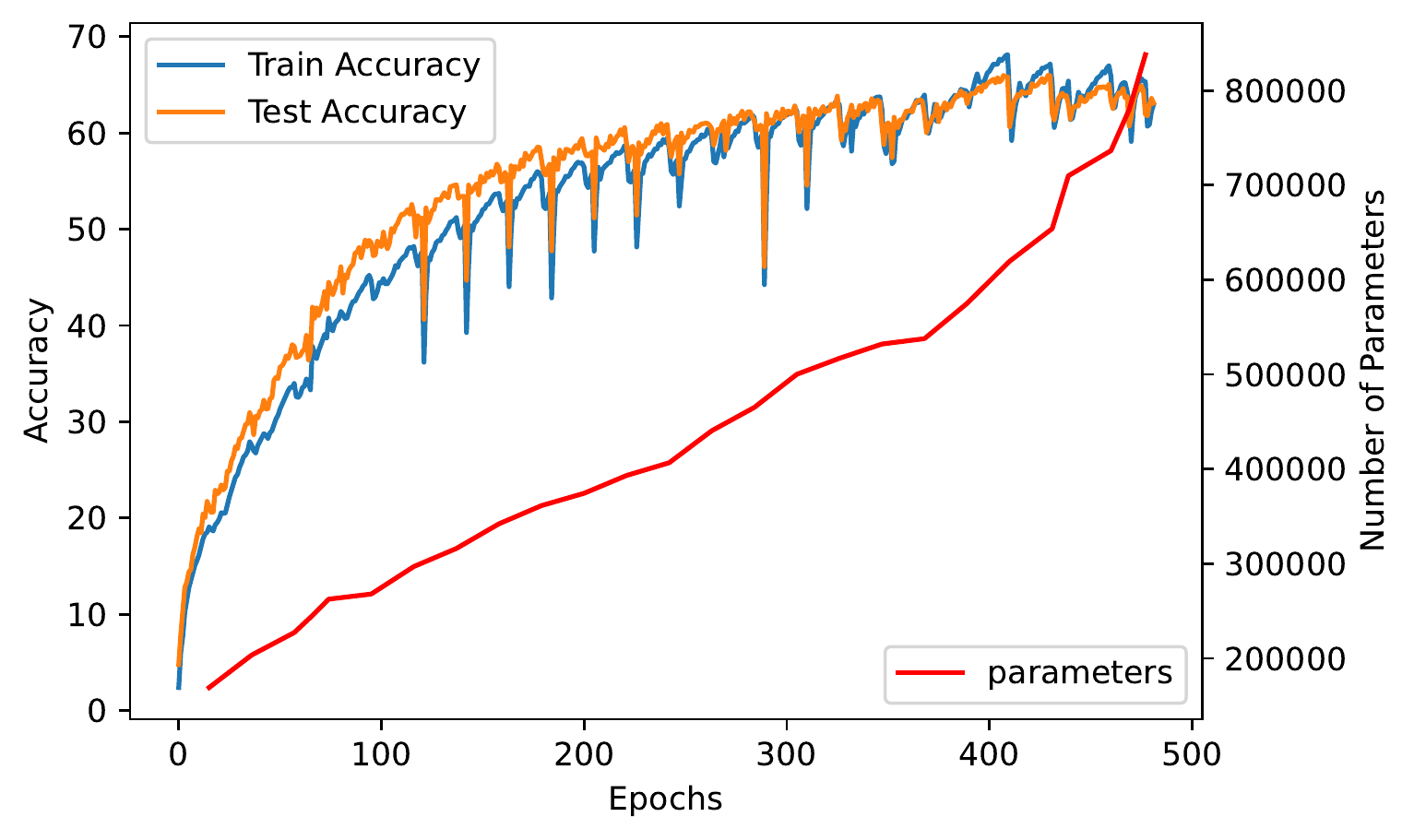}
     }
     \subfigure{
     \includegraphics[trim= 0.0cm 0.0cm 0.0cm 0.0cm, clip, width=0.7\textwidth]{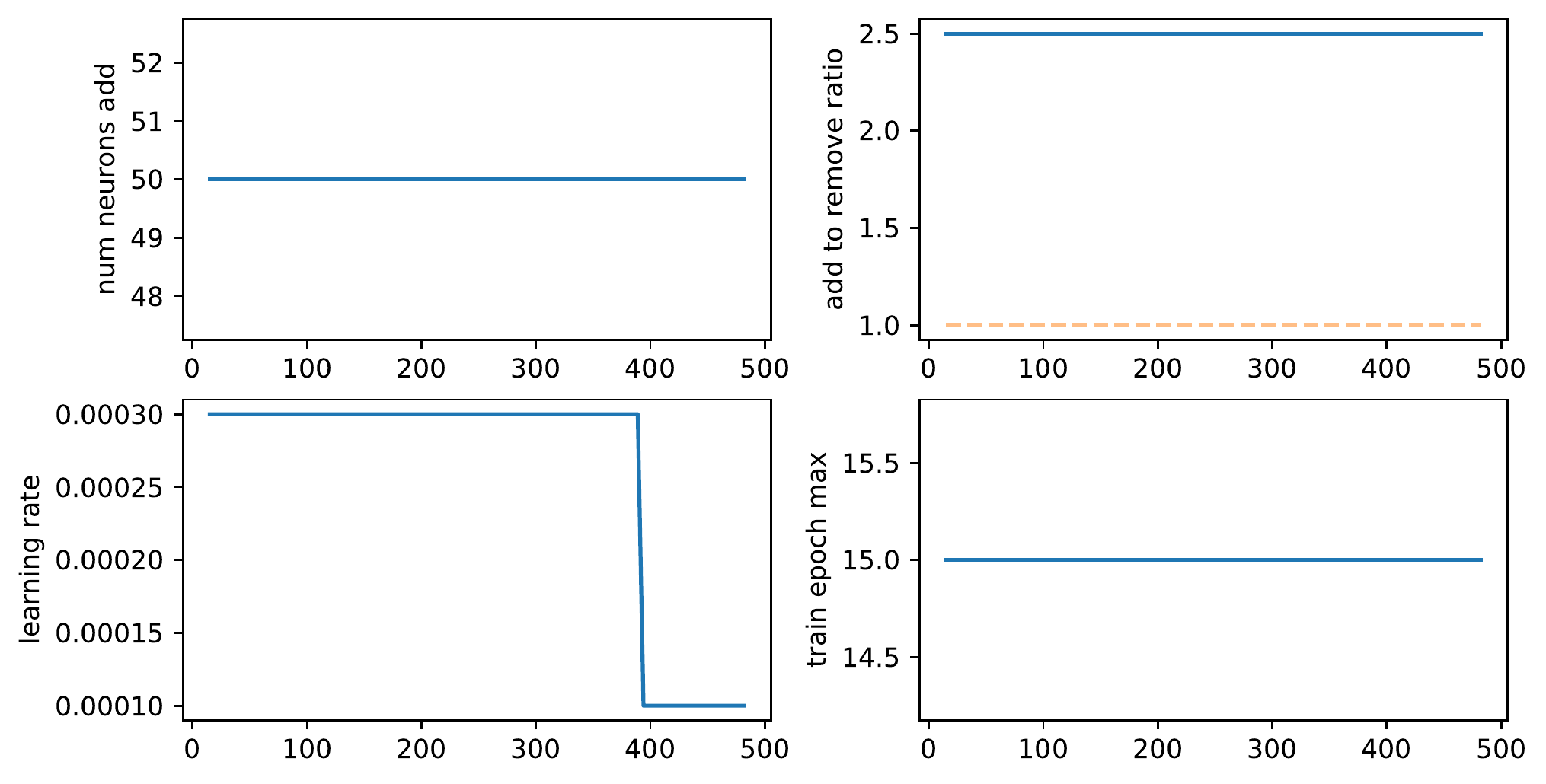}
     }
     \caption{\textbf{Top:} Epoch vs (Accuracy and Parameter) plot for CIFAR-100 dataset, \textbf{Bottom Four:} Meta-parameters used during CIFAR-100 training. We only modify the learning rate once during the search.}
     \label{fig:step_vs_param_acc_c100}
\end{figure}

This paragraph is extension of the CIFAR-100 experiment mentioned in the Experiments Section (\ref{sec:experiment}).
We also show the relationship between Epochs, Accuracy and Number of parameters in the Figure~\ref{fig:step_vs_param_acc_c100}. Furthermore, we also show the values of meta-parameters used in the experiment in the same Figure. In CIFAR-100 experiments, we find trends similar to experiments on CIFAR-10. Our method produces architecture having $65.98$\% accuracy with $0.654$M parameters whereas ResNet-18 with comparable parameters of $0.627$M has an accuracy of $65.43$\%.

\section{Limitations and Future Work}

We implement Convolutional Hierarchical Residual Network using only 3x3 convolution operations. Many NAS algorithms search for kernel size as well as strides and padding. We could add another variety of Convolution parameters in search space in future work. Furthermore, we can also add other search variables such as dropout rate and activation functions.

In future works, we could also use better neuron initialization methods such as GradMax~\cite{evci2022gradmax} to fit the residual better, and use better importance estimation algorithm for pruning as well. We can also implement better optimization techniques to avoid accuracy drops while training newly added neurons. Furthermore, we can apply meta models to learn the meta-parameters which in turn modifies the network. We also plan to optimize the Hierarchical Residual Architecture for better performance.

\end{document}